\documentclass[sigconf]{acmart}

\usepackage{booktabs} % For formal tables
\usepackage{soul}
\usepackage{color,soul} % highlight package

\acmDOI{10.475/123_4}

\acmISBN{123-4567-24-567/08/06}

\copyrightyear{2016}

\acmArticle{4}
\acmPrice{15.00}

\newif\ifsubmit
\submitfalse
\ifsubmit
    \newcommand{\aayush}[1]{}
    \newcommand{\todo}[1]{}
    \newcommand{\tocite}[1]{}
\else
    \newcommand{\aayush}[1]{[{\color{red}AA: #1}]}
    \newcommand{\todo}[1]{[{\color{red}TODO: #1}]}
    \newcommand{\tocite}[1]{[{\color{red}CITE: #1}]}
\fi

\begin{document}
\title[Pseudo Agent-Based Hyperparameter Optimization for Efficient Accelerator Design]{PABO: Pseudo Agent-Based Multi-Objective Bayesian Hyperparameter Optimization for Efficient Neural Accelerator Design}

\author{Maryam Parsa$^1$, Aayush Ankit$^1$, Amirkoushyar Ziabari$^{1,2}$, Kaushik Roy$^1$}
\affiliation{%
\institution{1. Purdue University}
}
\affiliation{%
\institution{2. Oak Ridge National Laboratory}
}
\email{{mparsa, aankit, kaushik}@purdue.edu}
\email{ziabariak@ornl.gov}
\renewcommand{\shortauthors}{M. Parsa, A. Ankit, A. Ziabari, K. Roy}

\begin{abstract}
%With the increase in the complexity of Deep Neural Networks' (DNNs) architectures, and the demand for energy efficient hardware platforms for adapting DNNs on low-power mobile devices, it is impractical to manually tune DNN and hardware hyperparameters in order to obtain maximum performance while minimizing the energy requirements of the underlying hardware.
The ever increasing computational cost of Deep Neural Networks (DNN) and the demand for energy efficient hardware for DNN acceleration has made accuracy and hardware cost co-optimization for DNNs tremendously important, especially for edge devices.
Owing to the large parameter space and cost of evaluating each parameter in the search space, manually tuning of DNN hyperparameters is impractical.
Automatic joint DNN and hardware hyperparameter optimization is indispensable for such problems. 
Bayesian optimization-based approaches have shown promising results for hyperparameter optimization of DNNs. 
%These techniques, however, developed either without considering the underlying hardware implementation resulting in inefficient designs; or few recent works that perform joint optimization of hyperparameters are not generalizable and mainly focused on CMOS-based architectures.
However, most of these techniques have been developed without considering the underlying hardware, thereby leading to inefficient designs.
Further, the few works that perform joint optimization are not generalizable and mainly focus on CMOS-based architectures.

In this work, we present a novel pseudo agent-based multi-objective hyperparameter optimization (PABO) for maximizing the DNN performance while obtaining low hardware cost. 
%Compared to existing methods, PABO poses a theoretically different approach for joint optimization of performance of a DNN with large design space implemented on a neuromorphic-based memristive crossbar accelerator.
Compared to the existing methods, our work poses a theoretically different approach for joint optimization of accuracy and hardware cost and focuses on memristive crossbar based accelerators.
%In this technique, a supervisor agent establishes a connection between posterior Gaussian distribution models of network accuracy and hardware energy requirements.
PABO uses a supervisor agent to establish connections between the posterior Gaussian distribution models of network accuracy and hardware cost requirements.
%The agent helps reducing the mathematical complexity of co-optimization problem by removing the unnecessary computation and updates of acquisition functions. 
%Thus, in turn it drastically speeds up the optimization procedure.
The agent reduces the mathematical complexity of the co-optimization problem by removing unnecessary computations and updates of acquisition functions, thereby achieving significant speed-ups for the optimization procedure. 
%The method outputs a Pareto frontier that shows a trade-off between designing a high-performance network and an energy efficient hardware.
PABO outputs a Pareto frontier that underscores the trade-offs between designing high-accuracy and hardware efficiency.
Our results demonstrate a superior performance compared to the state-of-the-art methods both in terms of accuracy and computational speed ($\sim$100x speed up).%$\times$

\end{abstract}

%
% The code below should be generated by the tool at
% http://dl.acm.org/ccs.cfm
% Please copy and paste the code instead of the example below.
%

\maketitle
%\vspace{-0.1cm}
%%%%%%%%%%%%%%%%%%%%%%%%%%%%%%%%%%%%%%%%%%%%%%%%%%%%%%%%%%%%%%%%%%%%%%%
%%%%%%%%%%%%%%%%%%%%%%%%%%%%%%%%%%%%%%%%%%%%%%%%%%%%%%%%%%%%%%%%%%%%%%%
%%%%%%%%%%%%%%%%%%%%%%%%%%%%%%%%%%%%%%%%%%%%%%%%%%%%%%%%%%%%%%%%%%%%%%%
%%%%%%%%%%%%%%%%%%%%%%%%%%%%%%%%%%%%%%%%%%%%%%%%%%%%%%%%%%%%%%%%%%%%%%%

\section{Introduction}

Advances in computing engines and graphic processing units (GPUs) as well as massively produced data from smart devices, social media, and internet, create an immense opportunity for machine learning and in particular deep learning techniques to solve brain-inspired tasks such as recognition and classification. 
Deep neural networks (DNNs) are computationally expensive, and require substantial resources.
Therefore, domain-specific energy-efficient accelerators built with CMOS~\cite{chen2016eyeriss, jouppi2017tpu, venkataramani2017scaledeep, fowers2018configurable}, resistive crossbars~\cite{shafiee2016isaac, chi2016prime, ankit2019puma} and spintronics~\cite{ramasubramanian2014spindle} based technologies were proposed to boost the performance and speed of DNNs. 

Performance of DNNs is highly dependent on the selection of hyperparameters (HPs). These include, but are not limited to, the number of hidden layers, kernel sizes, the choice of optimizer and non-linearity function. On the other hand, designing a neuromorphic hardware is reliant not only on a networks' HPs, but also on hardware specific parameters such as memory bandwidth, pipelining in CMOS technologies~\cite{reagen2017case}, and the number of bits, the number of crossbars and the crossbar sizes in memristive crossbar accelerators~\cite{ankit2019puma}. Over the past few years, there has been a growing interest in optimizing DNNs' HPs to obtain their optimum performance (in terms of accuracy, speed, etc). Most of these HP optimization techniques, however, have been developed with little or even no attention to the underlying hardware implementation. 

Optimizing the network performance without considering the strong correlation between its HPs and the corresponding hardware specific parameters results in a sub-optimal and inefficient hardware architecture.
In fact, designing an energy-efficient accelerator for implementing a high performance DNN (i.e. with minimum error) is a multi-objective optimization task with expensive black-box  objective functions, and solution to such optimization problems is a Pareto frontier.
The Pareto frontier is a set that consists of solutions in which no other is superior in optimizing both objective functions (minimizing the DNN's error and maximizing the hardware energy efficiency).
In other words, each member of the Pareto set is not dominated by other members of the set, where the dominance is defined as: The vector $\vec{a}$ dominates vector $\vec{b}$ notated as $\vec{a} \succ \vec{b}$ or  $\vec{b} \prec \vec{a}$, iff $ \forall{i};f_i{(a)} \leq f_i{(b)}$ where $f_i$ is the $i$-th objective function~\cite{okabe2003critical}. 
Each point in the Pareto set represents an HP combination and signifies a trade-off between the two objective functions (minimizing the DNN's error and maximizing the hardware energy efficiency). 
Designer may select any HP combination from the Pareto frontier set according to the design requirements, DNN accuracy, and hardware energy efficiency. 

Bayesian optimization is one of the most effective techniques to optimize HPs of a DNN~\cite{loshchilov2016cma}. This stochastic approach is an online optimization of an expensive objective function \textit{f}, when \textit{f} is unknown. Starting with a random selection of HPs from the search space (observation), a Gaussian process based posterior model is estimated. For each posterior model, an acquisition function is defined based on a trade-off between exploring the search space, and exploiting the current status. The optimum point of the acquisition function directs us to the next set of HPs to evaluate. The posterior model is updated with each new experiment until adding an extra observation does not improve the model~\cite{bergstra2011algorithms}.

\begin{figure}[h]
\centering
\includegraphics [scale=.25]{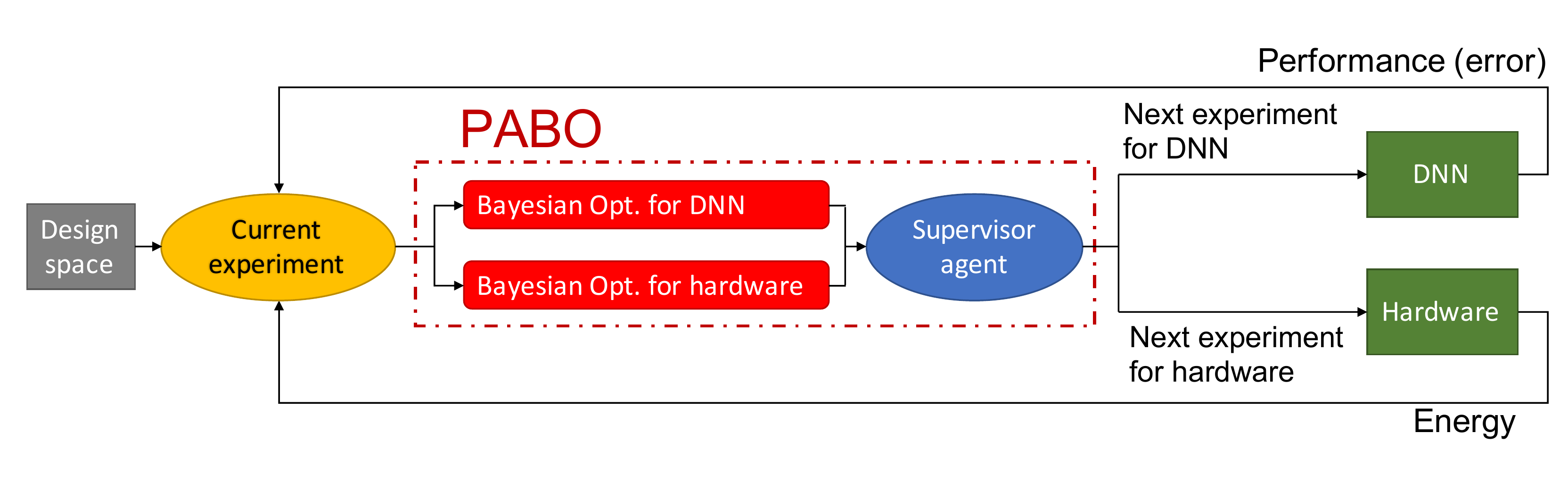}
\caption{
Overview of pseudo agent-based multi-objective Bayesian optimization (PABO) search process.  
}
%\vspace{-0.95cm}
\label{fig1} 
\end{figure}         

In this paper, we propose a pseudo agent-based multi-objective Bayesian optimization technique using Gaussian distribution (PABO). Figure~\ref{fig1} summarizes the PABO search process. 
The core of this approach is the PABO block which uses Bayesian optimization to estimate posterior models, and thus, find optimum HPs for the two black-box objective functions, related to DNN's performance and hardware's energy requirement.
This process is then followed by a supervisor agent that evaluates the impacts of output HPs on the other posterior model and decides which HPs it must pass along.
The PABO block decides on the set of HPs to evaluate at each step, the direction of the search process, and when to stop the technique. 
Using a supervisor agent in the PABO block reduces the complexity of the joint optimization problem, which in turn speeds up the algorithm to obtain the Pareto frontier compared to the state-of-the-art methods.
In fact, in all the experiments performed, the output Pareto frontier is calculated with few iterations.
Such capability is further beneficial for solving multi-objective problems with more than two objective functions.  
%For the underlying hardware, we selected a memristive crossbar accelerator due to its flexibility, scalability and energy efficiency ~\cite{ankit2019puma}. 

We made the following contributions:

\begin{enumerate}
	\item A novel pseudo agent-based multi-objective Bayesian optimization technique that estimates the Pareto frontier with only few evaluations of network training and hardware energy calculation. Further, the proposed multi-objective black-box optimization approach can be applied to any number of black-box functions. For example, user can optimize neural network accuracy, hardware energy efficiency, and area requirements simultaneously. 
    \item To the best of our knowledge, for the first time, a highly efficient memristive crossbar architecture~\cite{ankit2019puma} is selected to demonstrate the need of higher level optimization techniques for using neural networks. With memristive crossbar architectures, %\aayush{are we varying crossbar size?}
    we demonstrate significant enhancement in hardware energy requirements when the network and hardware HPs are optimized jointly.
    \item We compare PABO with state-of-the-art methods and show that it estimates the Pareto frontier with superior computational speed. 
    %\item \MP{The high speed of our algorithm allows us to consider the network architecture selection (i.e. AlexNet, GoogleNet, ResNet, ...) as a HP, and use PABO to find the optimized architecture for a specific hardware or dataset.}
\end{enumerate}

%%%%%%%%%%%%%%%%%%%%%%%%%%%%%%%%%%%%%%%%%%%%%%%%%%%%%%%%%%%%%%%%%%%%%%%
%%%%%%%%%%%%%%%%%%%%%%%%%%%%%%%%%%%%%%%%%%%%%%%%%%%%%%%%%%%%%%%%%%%%%%%
%%%%%%%%%%%%%%%%%%%%%%%%%%%%%%%%%%%%%%%%%%%%%%%%%%%%%%%%%%%%%%%%%%%%%%%
%%%%%%%%%%%%%%%%%%%%%%%%%%%%%%%%%%%%%%%%%%%%%%%%%%%%%%%%%%%%%%%%%%%%%%%
%\vspace{-0.25cm}
\section{Related Work}

In this section we provide a brief overview of the available multi-objective optimization techniques that have been used to co-optimize the neural network performance and hardware energy requirement.

The traditional approaches to solve an HP optimization problem, are grid search, random search and stochastic grid search~\cite{bergstra2011algorithms, larochelle2007empirical, bergstra2012random}. The grid search technique is a manual search that estimate the optimum hyperparameter (HP) set by training a DNN for all possible combinations of HPs. This approach is only usable in small networks with few HPs. As its names indicates, the random search technique performs a random search over some domain of the exhaustive search space. Although this approach is more practical compared to the grid search, it is not necessarily an efficient approach since it might search spaces that are not promising~\cite{smithson2016neural}. 
Stochastic grid search~\cite{bergstra2012random} is another traditional approach that adds stochasticity to the manual search method to reduce the number of required network training for different combinations of HPs. This technique is also inefficient as it blindly searches the space. 

Heuristic and sequential approaches such as simulated annealing (SA), genetic algorithm (GA), and Bayesian optimization techniques are more common to optimize a network HPs~\cite{snoek2012practical, hernandez2016general, hernandez2015predictive, hernandez2014predictive}. Among them GA and Bayesian approaches are favorable. A key limitation of the SA is that it might trap in a local minima for complex problems such as DNNs. GA technique speeds up the process of search for the next HP combination by isolating the evaluated HP sets with good network performance and build the next HP set based upon those good genes. Bayesian approaches are leveraging the stochasticity of Gaussian processes and conditional probability to efficiently limit the search space to HP sets that produce the best network performance.

More recently, for multi-objective optimization problems in which the objective functions are expensive to evaluate, design space exploration (DSE)~\cite{domhan2015speeding, smithson2016neural}, non-dominated sorting genetic algorithm (NSGA-II)~\cite{deb2002fast}, evolutionary optimization~\cite{michel2019dvolver}, and Bayesian optimization techniques ~\cite{reagen2017case, bradford2018efficient, hernandez2016predictive, hernandez2016designing} were developed. 
DSE~\cite{smithson2016neural} uses deep neural networks to find the optimum HP set for the best network performance while designing an energy-efficient hardware. However, designing such a DNN is application dependent and it cannot efficiently be generalized to other problems.
NSGA-II algorithm is a modified GA based search technique for multi-objective optimization problems with expensive black-box functions. Evolutionary processes such as genetic crossover and mutation is utilized to guide the population toward the Pareto frontier~\cite{deb2002fast}. We used this technique as a benchmark to compare with the PABO results. Both techniques estimate Pareto frontier, but the PABO is $\sim$100x faster.

A thorough review on Neural Architecture Search (NAS) for multi-objective optimization is given at \cite{michel2019dvolver}. Different approaches such as reinforcement learning based for NAS problems, evolutionary algorithm based, and search acceleration were among the techniques reviewed in \cite{michel2019dvolver} with the main focus on the most recent algorithms of MONAS~\cite{hsu2018monas} and DPP-Net~\cite{dong2018dpp}. The experimental results show that these approaches are able to find Pareto fronts with respect to different objectives imposed by devices and networks' performances. However, since these techniques are all based on evolutionary optimization, they require substantial computational resources. In PABO we use fundamentally different optimization technique (Bayesian-based) and able to find Pareto frontier set much faster than evolutionary based approaches.

Bayesian optimization based techniques were already used in the literature for optimizing HPs~\cite{reagen2017case, bradford2018efficient, hernandez2016predictive}. Reagen in \cite{reagen2017case} designed Spearmint package that outputs the Pareto frontier for a design space that minimizes energy consumption in Aladdin accelerator and optimizes the DeepNet performance. While promising results are demonstrated using this technique, the Spearmint package is not readily available to be used with the other networks and accelerator architectures. In addition, the method cannot be extended to more than two objective functions.

Inspired by the success of the Bayesian optimization techniques and to serve our purpose, which is optimizing HPs for a memristive crossbar accelerator~\cite{ankit2019puma}, we designed a novel agent-based Bayesian optimization algorithm to find the optimum set of HPs that maximize the DNN performance and energy efficiency of the underlying hardware. 
%\aayush{I think we should not distinguish babsed on past works not doing real hardware and we doing it. Hardware architectures are prone to change and hence, generality to abstract hardware may be more important. I feel discussing we focussing on memritive accelerator and past works on CMOS is more important point. We also need to justify why an algorithm for CMOS may not work as well for memristive hardware}
The proposed hardware-aware optimization method is not limited to the number of functions to optimize (network performance, training speed, hardware area requirements or energy consumption) %\aayush{add what functions are/maybe of interest}
, and it is generic enough that can be used with different neural networks and accelerator architecture designs.

%The \textit{proceedings} are the records of a conference.\footnote{This is a footnote}

%%%%%%%%%%%%%%%%%%%%%%%%%%%%%%%%%%%%%%%%%%%%%%%%%%%%%%%%%%%%%%%%%%%%%%%
%%%%%%%%%%%%%%%%%%%%%%%%%%%%%%%%%%%%%%%%%%%%%%%%%%%%%%%%%%%%%%%%%%%%%%%
%%%%%%%%%%%%%%%%%%%%%%%%%%%%%%%%%%%%%%%%%%%%%%%%%%%%%%%%%%%%%%%%%%%%%%%
%%%%%%%%%%%%%%%%%%%%%%%%%%%%%%%%%%%%%%%%%%%%%%%%%%%%%%%%%%%%%%%%%%%%%%%
%\vspace{-0.2cm}
\section{Background}\label{sec3}

% ref: CNN for dummies papers\MOO\10.1.1.727.1832.pdf
In this section, we provide some background on the Bayesian optimization and in particular Gaussian processes with acquisition functions.

%%%%%%%%%%%%%%%%%%%%%%%%%%%%%%%%%%%%%%%%%%%%%%%%%%%%%%%%%%%%%%%%%%%%%%%
%\subsection{Bayesian Optimization: The Gaussian Process Approach}\label{ssec31}

Bayesian optimization (BO) is a derivative-free sequential optimization technique which is widely used when the objective function $f$ is black-box and/or expensive to evaluate. 
One of the main variants of the BO is the Gaussian processes (GP) with acquisition function (AF).
A GP is a stochastic process in which each combination of set of random variables follows a multivariate Gaussian distribution and can be defined by its mean and variance~\cite{bergstra2011algorithms}. A key property they hold is that, the conditional distribution, and in turn the corresponding posterior distribution, of an expensive black-box function $f$ at finite number sampling points will be a GP, if the prior distribution of it is assumed to be a GP~\cite{bergstra2011algorithms}.

\begin{figure}[h]
\centering
\includegraphics [scale=.26]{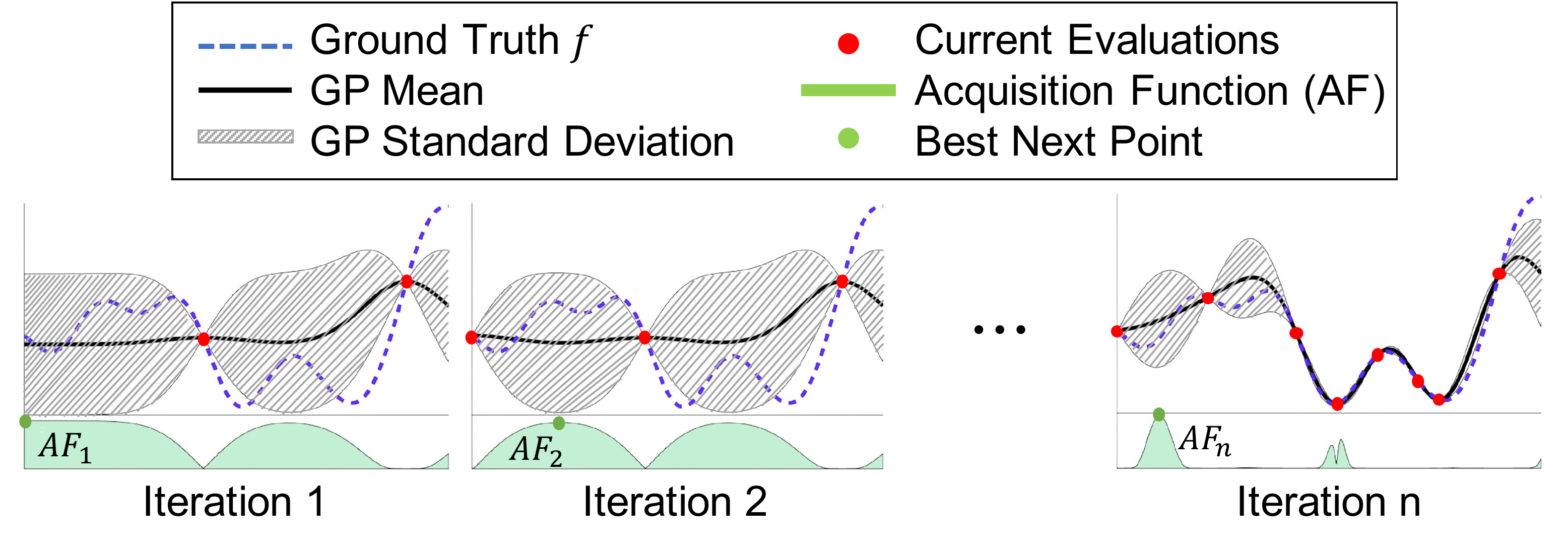}
\caption{
An overview of the Bayesian optimization using Gaussian Process (GP) with Acquisition Function (AF) method.
The figure is slightly modified from its original in~\cite{Adams2014}.}
%\vspace{-0.4cm}
\label{fig2} 
\end{figure}  

Figure~\ref{fig2} shows the procedure to estimate $f$ using the GP with AF. Here, we would like to estimate the blue dashed curve that is the ground truth for the unknown function $f$. We assume that the observation $y_i$ is the output of function $f$ for the input $\theta_i$. For all current observations of $D_n=\{(\theta_1,y_1),...,(\theta_n,y_n)\}$, we implement BO to predict the best next point $\theta_{n+1}$ to evaluate $f$ conditioned on $D_n$. This predictive conditional distribution is called posterior Gaussian model, and it is calculated by conditional probability of $p(y_{n+1}|\theta_{n+1},D_n)$. In the Figure~\ref{fig2}, mean and standard deviation of posterior Gaussian model is shown in solid black line and the shaded area, respectively. This model is then used to calculate $AF$, which directs the BO to the best next observation by exploiting and exploring the design space. Exploitation is seeking for places with the lowest mean (to stay near high quality observations), and exploration is to explore the search space where the variance is high (to cover the unknown regions and avoid trapping in local optima). AF is computed by the expected utility of evaluating function $f$, and it is a way to predict how useful it is to explore $\theta_{n+1}$ in order to optimize the black-box function $f$ ~\cite{reagen2017case}. 

%\begin{equation}
%    AF(x) = E_{p(y_{n+1}|\theta,D_n)}[\alpha(y_{n+1}|x,D_n)]
%\label{AF}    
%\end{equation}}

The acquisition function is shown in green in Figure~\ref{fig2}. Optimum point of the AF shows the best next HP set to evaluate in the next iteration $(\theta_{n+1})$. The posterior Gaussian distribution model is updated accordingly with the value of function $f$ at $\theta_{n+1}$. A new surrogate acquisition function is calculated for the updated posterior Gaussian model to obtain the next evaluation point.
We observe that as we evaluate more points, the variance of the posterior Gaussian model is reduced, and the estimated mean gets closer to the ground truth function $f$ that we would like to approximate.
This procedure iteratively continues until the next point derived from optimizing the updated acquisition function cannot change to the estimated mean and variance of the posterior Gaussian model. 

In PABO, we use Bayesian optimization processes along with a supervisor agent as the core processing unit to calculate the Pareto frontier of the multi-objective optimization of maximizing the DNN performance while maintaining the hardware energy requirements minimum.
Details of our method is given in the following section.

%%%%%%%%%%%%%%%%%%%%%%%%%%%%%%%%%%%%%%%%%%%%%%%%%%%%%%%%%%%%%%%%%%%%%%%
%%%%%%%%%%%%%%%%%%%%%%%%%%%%%%%%%%%%%%%%%%%%%%%%%%%%%%%%%%%%%%%%%%%%%%%
%%%%%%%%%%%%%%%%%%%%%%%%%%%%%%%%%%%%%%%%%%%%%%%%%%%%%%%%%%%%%%%%%%%%%%%
%%%%%%%%%%%%%%%%%%%%%%%%%%%%%%%%%%%%%%%%%%%%%%%%%%%%%%%%%%%%%%%%%%%%%%%
\section{Methodology}

The overview of the proposed PABO method was given in Figure~\ref{fig1}. In this section we explain this technique in more details. 
Algorithm~1 shows the PABO algorithm flow. 
We assume \textbf{\textit{err}} (DNN's performance) and \textbf{\textit{eng}} (hardware's energy requirement) are the two expensive black-box functions we would like to optimize. 

The PABO process can be explained as follows. PABO has two key elements, the BO processes and the supervisor agent.
In the $n^{th}$ iteration, two BO processes are run. The first estimated BO with Gaussian processes computes an acquisition function to find the optimum HPs, $\theta_{n+1}$ set, for an unknown \textbf{\textit{err}} function. The second estimated BO, on the other hand, calculates an acquisition function to find the optimum HPs, $\Gamma_{n+1}$ set, for an unknown \textbf{\textit{eng}} function.
The sets $D_{err}$ and $D_{eng}$ keep track of $\theta_{n+1}$ and $\Gamma_{n+1}$, respectively, so that they can be used as training data in the BO processes of the next iteration.
Before, running the next iteration, a supervisor agent is used to link these two seemingly separate processes. 
The supervisor agent evaluates sets $[err(\theta_{n+1}), eng(\theta_{n+1})]$, and $[err(\Gamma_{n+1}), eng(\Gamma_{n+1})]$, and examines whether these sets are dominated by all previous observations ($[err(\theta_{n}), eng(\theta_{n})]_{n=1}^{n}$ or $[err(\Gamma_{n}), eng(\Gamma_{n})]_{n=1}^{n}$) or not. The non-dominated sets may belong to the Pareto frontier of the problem and the corresponding HP set ($\theta_{n+1}$, or $\Gamma_{n+1}$) will be added to $D_{eng}$, or $D_{err}$, respectively.
Hence, the BO optimization processes of the next iteration has a more complete training set to estimate the acquisition functions and in turn optimized HPs in the direction of optimizing both objective functions. 
In what follows we explain the entire process in more details.

\begin{figure}[h]
\centering
\includegraphics [scale=.25]{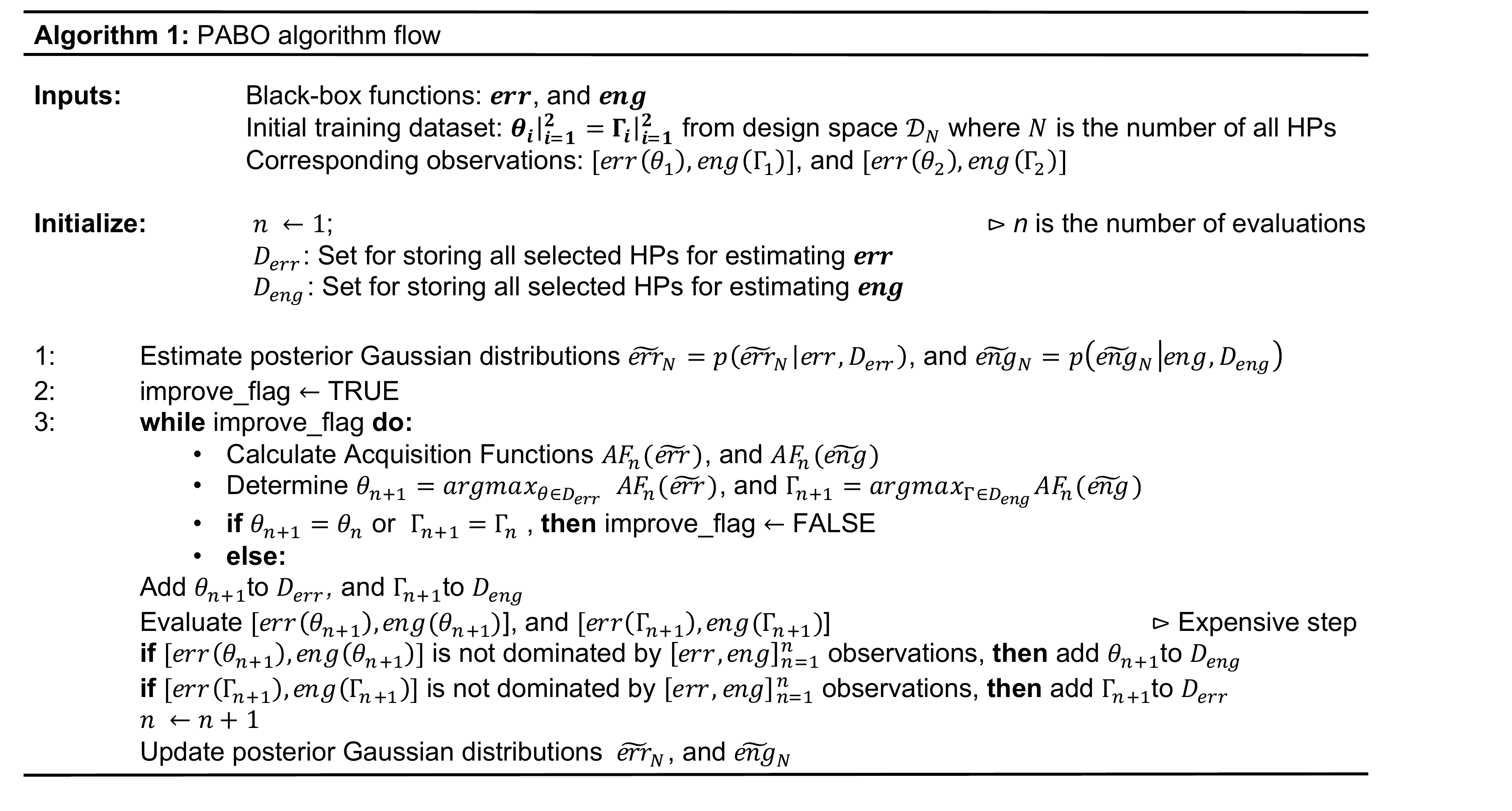}
%\vspace{-0.4cm}
%\label{al1} 
\end{figure} 

\hspace{-0.38cm}\textbf{Bayesian Optimization (BO) with Gaussian Processes (GP):}
In each iteration, BO finds the optimum HP sets that optimize the unknown \textbf{\textit{err}} and \textbf{\textit{eng}} objective functions, respectively.
To start the BO process, we need to have at least two initial training dataset with their corresponding set of observations. $\theta_1=\Gamma_1$ and $\theta_2=\Gamma_2$ are randomly selected HP sets from design space $D_N$ with observations of $[err(\theta_1), eng(\Gamma_1)]$, and $[err(\theta_2), eng(\Gamma_2)]$. 
For $N$ possible HP sets, $D_N$ is the exhaustive search space (grid search) that includes all HP combinations.
$D_{err}$, and $D_{eng}$ initially store the selected sets of HPs to optimize \textbf{\textit{err}} and \textbf{\textit{eng}} objective functions, respectively. 
They get updated after each iteration based on the supervisor agent's decision described later. 

As mentioned in section~\ref{sec3}, for each objective function, we estimate a posterior Gaussian model (shown with black line and shaded grey area in Figure~\ref{fig2}). In the $n-th$ evaluation, these predictive conditional distributions are defined by 
$p(\widetilde{err}_N|err_n,D_{err})$ and $p(\widetilde{eng}_N|eng_n,D_{eng})$ for \textbf{\textit{err}} and \textbf{\textit{eng}} objective functions, respectively. The former is the conditional probability of the approximated error function, given the current estimate of error values for the input HP sets. Similarly,  the latter is the conditional probability of the the approximated energy function, given the current estimate of energy values for the input HP sets. We calculate separate acquisition functions for each posterior Gaussian model. Details of defining this function is given in section~\ref{sec3}. The optimum points of the acquisition functions, are the best selected HP sets to evaluate in the next iteration ($\theta_{n+1}$, and $\Gamma_{n+1}$).

\hspace{-0.38cm}\textbf{Supervisor Agent:} 
Before running the next iteration, the next HP sets ($\theta_{n+1}$, and $\Gamma_{n+1}$) are input to a supervisor agent. $\theta_{n+1}$ is the new HP combination chosen by the BO algorithm for estimating \textbf{\textit{err}} function and will be added to $D_{err}$ set. Similarly, $\Gamma_{n+1}$, which is chosen by the BO algorithm for estimating \textbf{\textit{eng}} function, will be added to $D_{eng}$ set.

In addition, $\Gamma_{n+1}$ (the next selected HP set for optimizing the \textbf{\textit{eng}} function) will also be added to $D_{err}$ set if and only if the set $[err(\Gamma_{n+1}),eng(\Gamma_{n+1})]$ is not dominated by any other previous observations and might belong to the Pareto frontier set. This means that although $\Gamma_{n+1}$ is chosen by posterior Gaussian model and its corresponding AF for \textbf{\textit{eng}} function, it is also creating a solution set of $[err(\Gamma_{n+1}), eng(\Gamma_{n+1})]$ that up to this point belongs to the Pareto set. This set might be substituted in the future evaluations if the new set creates superior solutions in at least one of the objective functions. Moreover, $\theta_{n+1}$ will also be added to $D_{eng}$ set if and only if the set $[err(\theta_{n+1}),eng(\theta_{n+1})]$ is not dominated by any other previous observation and might belong to the Pareto frontier set. We repeat this process until the acquisition function vanishes.

Throughout the process, the supervisor agent guides the search process to the Pareto frontier region and speeds up the procedure without adding extra complexity to the underlying mathematics. 

%%%%%%%%%%%%%%%%%%%%%%%%%%%%%%%%%%%%%%%%%%%%%%%%%%%%%%%%%%%%%%%%%%%%%%%
%%%%%%%%%%%%%%%%%%%%%%%%%%%%%%%%%%%%%%%%%%%%%%%%%%%%%%%%%%%%%%%%%%%%%%%
%%%%%%%%%%%%%%%%%%%%%%%%%%%%%%%%%%%%%%%%%%%%%%%%%%%%%%%%%%%%%%%%%%%%%%%
%%%%%%%%%%%%%%%%%%%%%%%%%%%%%%%%%%%%%%%%%%%%%%%%%%%%%%%%%%%%%%%%%%%%%%%

\section{Baseline Accelerator}
In this work, the underlying hardware of choice is a neural accelerator. In the following, we overview the hardware architecture, and explain the energy consumption computation for a memristive crossbar accelerator. %Since PABO is a general optimization framework, it can be applied to any software/hardware co-optimization platform. Therefore, only abstract architecture and energy consumption model is introduced. Details of the hardware design is given in}~\cite{ankit2019puma}.

\subsection{Accelerator Overview}

\begin{figure}[h] 
%\vspace{-0.2cm}
\centering
\includegraphics [scale=.35]{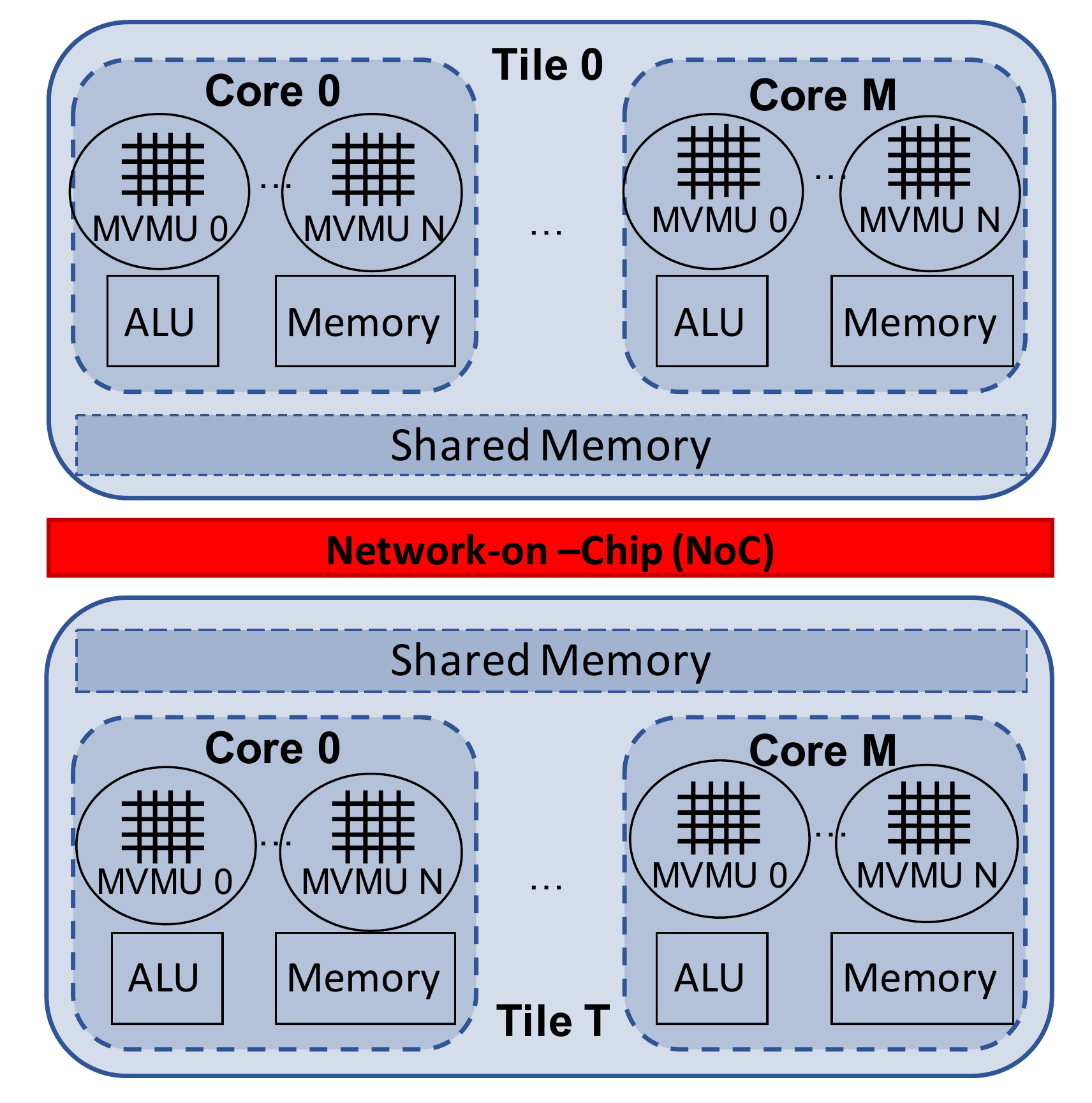}
%\vspace{-0.3cm}
\caption{
High-level Overview of a Abstract Hybrid Accelerator Architecture
}
%\vspace{-0.6cm}
\label{fig_arch} 
\end{figure}

This section provides an overview of the memristive crossbar based accelerator used in our evaluations.
Figure~\ref{fig_arch} shows a memristive crossbar based DNN accelerator.
Typical memristive accelerators employ a spatial architecture, where the DNN is executed by mapping the model across the on-chip crossbar storage in a spatial manner~\cite{ankit2019puma}.
This is because the memristive devices have high storage density, but are limited by the high write cost.
Consequently, the high storage density enables mapping DNNs spatially in practical die sizes while alleviating the high write cost which would be required if a crossbar was reused for different parts of the model in a time-multiplexed fashion.
At the lowest level, $N$ Matrix Vector Multiplication Units (MVMU) are grouped into a single core. 
Each MVMU is composed of multiple crossbars and performs a 16-bit $128 \times 128$ matrix-vector multiplication.
Note that multiple crossbars are needed to store high precision data required for DNN inference, since typical memristive crossbars store low-precision data such as 2-bits~\cite{shafiee2016isaac, ankit2019puma}.
At the next level, $M$ cores are grouped into a single tile with access to a shared memory, which enables data movement between cores (inter and intra tile). 
At the highest level, $T$ tiles are connected via a network-on-chip that enables data movement between tiles within a single node.
For large scale applications, multiple nodes can be connected using suitable chip-to-chip interconnect.

\subsection{Energy consumption}

We use an abstract energy consumption model to evaluate the efficiency for PABO, where we consider the energy consumption of the MVMUs only.
First, the abstract model enables evaluating the impact of hyperparameter optimization while isolating the benefits obtained from microarchitectural techniques.
This isolation enables widespread applicability where DNNs optimized with PABO can be executed over a wide range of memristive accelerators, where each accelerator may be leveraging different dataflow, compute to control granularity etc. 
Second, while a typical memristive accelerator expends significant energy in shared memory, network on chip and chip-chip interconnect due to the data movements in a spatial architecture, reducing the number of MVMU operations typically reduces the total energy consumption commensurately~\cite{ankit2017trannsformer}.

A layer (fully connected layer or convolution layer) is partitioned into smaller blocks of size N$\times$N to fit a MVMU (sized N$\times$N).
Each layer will map across multiple MVMUs that may span multiple cores and multiple tiles (see Figure ~\ref{fig_arch}).
Further, a MVMU may be used multiple times (once) for an input in a convolution layer (fully connected layer) due to weight-sharing.
Hence, the number of MVMU operations required to execute an inference of deep neural network will depend on the several HPs such as the number of layers, the number of extracted feature in each convolution layers, and the kernel sizes in the network architecture (Equations~\ref{eq_NC}, and ~\ref{eq_NF}).
%\vspace{-0.2cm}
\begin{equation}
%\vspace{-0.3cm}
    num\_xbar\_c_i = d_i \times d_i \times \lceil\dfrac{nc_i\times k_i \times k_i}{xs}\rceil \times \lceil\dfrac{nc_{i+1}}{xs}\rceil 
\label{eq_NC}    
\end{equation}

\begin{equation}
    num\_xbar\_f_i = \lceil\dfrac{nf_i}{xs}\rceil \times \lceil\dfrac{nf_{i+1}}{xs}\rceil
\label{eq_NF}    
\end{equation}

In these equations, $num\_xbar\_c_i$, and $num\_xbar\_f_i$ are number of crossbars for the $i_{th}$ convolution layer and the fully connected layers, respectively. $d_i$ is the dimension of the output, $nc_i$ is the number of input features for convolution layer $i$. Similarly, $nf_i$ is the number of input features for the fully connected layer $i$.
The term $d_i$ in equation~\ref{eq_NC} is for inherent weight-sharing property of convolution layers.

Typically, each memristive operation is followed by a vector linear, vector non-linear and data movement operations ~\cite{ankit2019puma}.
Consequently, the number of MVMU operations is proportional to the overall energy consumption and can be used as a metric of computational cost on hardware.
We calculate the total energy consumption in each convolution and fully connected layer based on the number of crossbar operations using the following equations.
In our selected memristive crossbar accelerator, a 16-bit (inputs and weights) crossbar operation (size $128$$\times$$128$) consumes ~$\simeq$44nJ energy. $epx$ is the energy per matrix vector multiplication operation.
The sum of energy consumption for all the convolution and fully connected layers is then used to calculate the total energy consumption of the memristive crossbar accelerator (Equation~\ref{eq_EngT}).

%\vspace{-0.35cm}
\begin{equation}
    tot\_eng\_t = (\sum_{i} num\_xbar\_c_i  + \sum_{i} num\_xbar\_f_i) \times  epx
\label{eq_EngT}    
\end{equation} 

 For each set of HPs given in Table~\ref{Tab1} case studies, we used Equations~\ref{eq_NC} through ~\ref{eq_EngT} to calculate the total hardware energy consumption. 
%%%%%%%%%%%%%%%%%%%%%%%%%%%%%%%%%%%%%%%%%%%%%%%%%%%%%%%%%%%%%%%%%%%%%%%
%%%%%%%%%%%%%%%%%%%%%%%%%%%%%%%%%%%%%%%%%%%%%%%%%%%%%%%%%%%%%%%%%%%%%%%
%%%%%%%%%%%%%%%%%%%%%%%%%%%%%%%%%%%%%%%%%%%%%%%%%%%%%%%%%%%%%%%%%%%%%%%
%%%%%%%%%%%%%%%%%%%%%%%%%%%%%%%%%%%%%%%%%%%%%%%%%%%%%%%%%%%%%%%%%%%%%%%

\section{Results}

To validate the proposed PABO algorithm, we used two different neural network architectures, AlexNet~\cite{krizhevsky2012imagenet}, and VGG19~\cite{simonyan2014very}, with flower17~\cite{nilsback2006visual} and cifar-10~\cite{krizhevsky2009learning} dataset, respectively. 
As the hardware of choice we selected a memristive crossbar accelerator~\cite{ankit2019puma}. A summary of the HPs that were selected for different case studies is given in Table~\ref{Tab1}. These choices were only made as the proof of concept for PABO technique. 
PABO search algorithm is not limited to these choices and depending on designer's decision, any type of neural network architecture, hardware, dataset, number and type of HPs can be selected.

In this section after discussing the limitations of single objective optimization, we present and analyze joint optimization results for different case studies in Table~\ref{Tab1}.

\begin{table}[]
\centering
\caption{Experimental Parameters for AlexNet and VGG19 Case Studies}
%\vspace{-0.1cm}
\label{Tab1}
\resizebox{\columnwidth}{!}{%}
\begin{tabular}{|c|c|c|c|c|}%{llll}
\hline
\textbf{Hyperparameter} & \textbf{Case Study 1} & \textbf{Case Study 2} & \textbf{Case Study 3} & \textbf{Type} \\ \hline 
\textbf{Architecture} & AlexNet & AlexNet & VGG19 & -\\ \hline \hline
\textbf{Dropout} & 0.4, 0.5 & 5e-3, ..., 5e-1 & - & DNN\\ \hline
\textbf{Learning rate} & 0.001 & 1e-6, ..., 1e-1 & 0.01, 0.1 & DNN\\ \hline
\textbf{Dropout, layer 1} & - & - & 0.3, 0.4 & DNN\\ \hline
\textbf{Learning rate decay} & - & - & 1e-6, 1e-4 & DNN\\ \hline
\textbf{Weight decay} & - & - & 0.05, 0.0005 & DNN\\ \hline
\textbf{Momentum} & 0.85, 0.9, 0.95 & 7e-3, .., 7e-1 & - & DNN \\ \hline
\textbf{\# of FC layers} & 2, 3 & 2, 3 & - & DNN/HW \\ \hline
\textbf{\# of Conv. layers} & 4, 5 & 4, 5 & - & DNN/HW\\ \hline
\textbf{Kernel size, layer 1} & 7, 5 & 3, 5, 7, 11 & - & DNN/HW\\ \hline
\textbf{Kernel size, layer 2} & 3, 5 & 3, 5 & - & DNN/HW\\ \hline
\textbf{Kernel Size, layer 3} & 3, 5 & 3, 5 & - & DNN/HW\\ \hline
\textbf{Kernel Size, layer 4} & 3 & 3, 5 & - & DNN/HW\\ \hline 
\textbf{Kernel size, layer 6} & - & - & 3, 5 & DNN/HW\\ \hline
\textbf{Kernel size, layer 7} & - & - & 3, 5 & DNN/HW\\ \hline
\textbf{Kernel Size, layer 8} & - & - & 3, 5 & DNN/HW\\ \hline
\textbf{Kernel Size, layer 9} & - & - & 3, 5, 7 & DNN/HW\\ \hline 
\textbf{\# of features, layer 1} & - & - & 64, 128 & DNN/HW\\ \hline       
\textbf{\# of features, layer 2} & - & - & 128, 256 & DNN/HW\\ \hline       
\textbf{\# of features, layer 4} & - & - & 256, 512 & DNN/HW\\ \hline     \hline   

\textbf{Search Space Size} & 192 & 6912 & 3072 & -\\ \hline \hline     

\end{tabular}
}
%\vspace{-0.5cm}
\end{table}

\subsection{Single-Objective Optimization}

\begin{figure}[h]
%\vspace{-0.3cm}
\centering
\includegraphics [scale=.26]{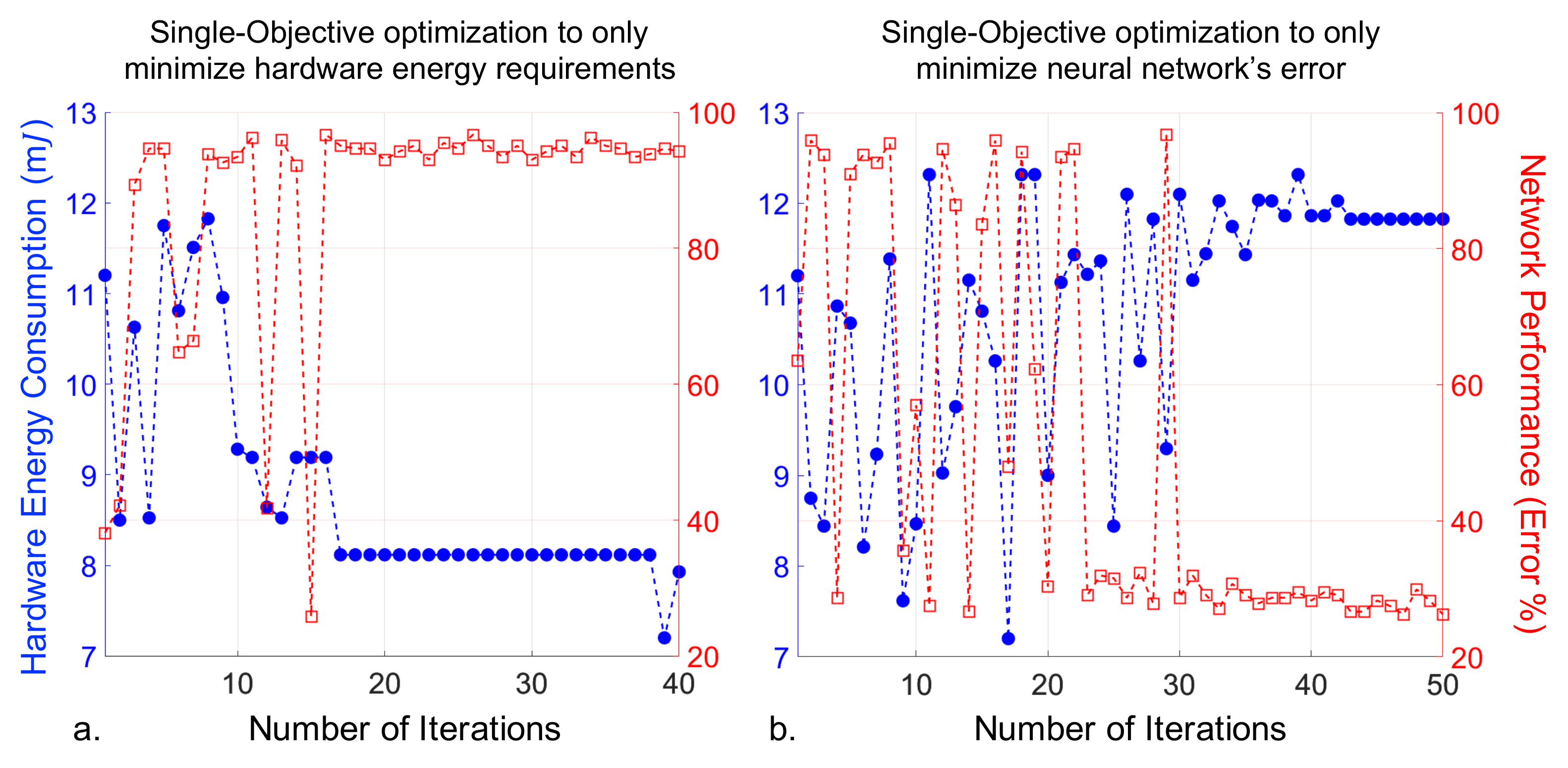}
%\vspace{-0.4cm}
\caption{
Single-objective optimization results for Table~\ref{Tab1}, case study 2 with SKOPT python Bayesian optimization package. 
a. Optimizing HPs for hardware energy consumption only. b. Optimizing HPs for DNN's performance only.
}
%\vspace{-0.3cm}
\label{fig4} 
\end{figure}

Before presenting the joint optimization results, it is imperative to answer the question: Why one cannot rely on single objective optimization to minimize the hardware energy consumption with maximum neural network accuracy? Figure~\ref{fig4} demonstrates the limitations of using independent single objective HP optimization techniques to separately design a neural network with optimum performance and a hardware with minimum energy requirements.

Figure~\ref{fig4}a shows that designing an energy efficient hardware without optimizing the network accuracy leads to significant decrease in the network performance (large error). The selected HP set is reported after 40 evaluations of hardware energy consumption. For this HP, the minimum energy is $\sim$7.8mJ, while the DNN's error is $\sim$92\%. Similarly, in Figure~\ref{fig4}b the network performance -in terms of reducing error- is optimized without considering the energy consumption of the underlying hardware. The inefficient hardware design is evident as the reported minimum error region occurs at high hardware energy consumption. Both of these results are undesirable, and are the main reasons to seek a multi-objective approach to find HPs that minimizes DNN's error while designing an energy-efficient hardware. We used SKOPT python package to solve these single-objective Bayesian optimization for AlexNet on Flower17 dataset with HPs given in Table~\ref{Tab1}, case study 2.

\subsection{Multi-Objective Optimization (PABO)}

\begin{figure}[h]
\centering
\includegraphics [scale=.44]{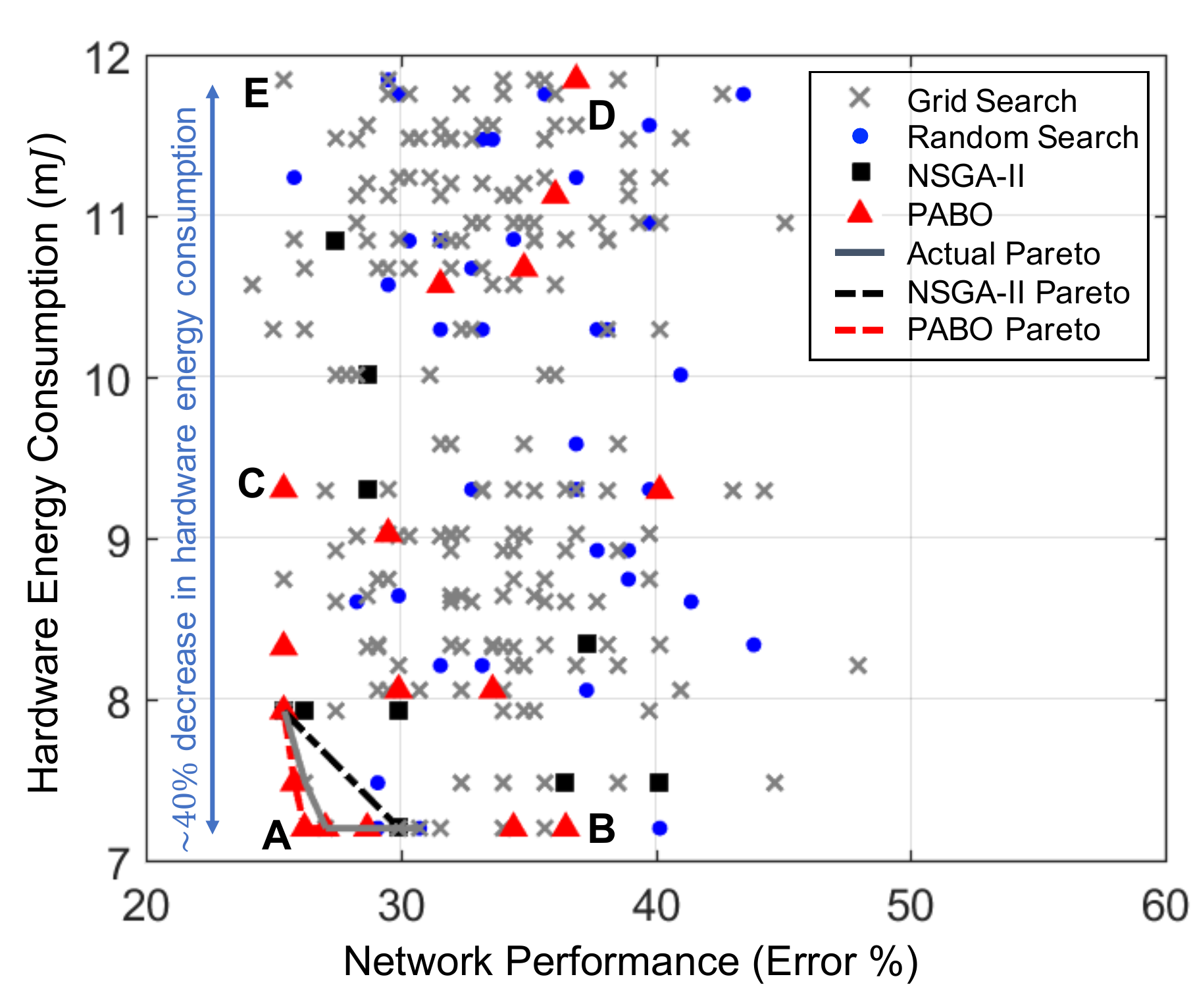}
%\vspace{-0.2cm}
\caption{
Case study 1: AlexNet on Flower17 dataset with 192 possible set of HPs. Comparison between grid search for all HP combinations (grey cross), random search with evaluating 40 different sets of HPs (blue dots), NSGA-II with population size of 10 and maximum generation of 50 (black squares), and PABO (red triangles).
The red dashed line, gray line and the black dashed line are the Pareto frontiers obtained by PABO, grid search and NSGA-II approaches.
}
%\vspace{-0.3cm}
\label{fig5} 
\end{figure}

We used the proposed PABO algorithm to find the optimum DNN accuracy while minimizing the underlying memristive crossbar accelerator energy consumption on three case study.
In case study 1, we used AlexNet network with the Flower17 dataset with a small search space of 192 HPs. The range of HPs are provided in Table 1.
In this case, we intentionally selected a small search space, so that we can estimate the actual Pareto frontier using the grid search method. 
Figure~\ref{fig5} shows the results for case study 1.
In this Figure, PABO's result compared with grid search, random search, and state-of-the-art NSGA-II (Non-Dominated Sorting Genetic Algorithm)~\cite{deb2002fast}.
Red triangles, blue dots, black squares and gray crosses correspond to PABO, random search, NSGA-II, and grid search, respectively.
Each point in the figure corresponds to one evaluation of the noted techniques.

\begin{figure}[h]
%\vspace{-0.2cm}
\centering
\includegraphics [scale=.6]{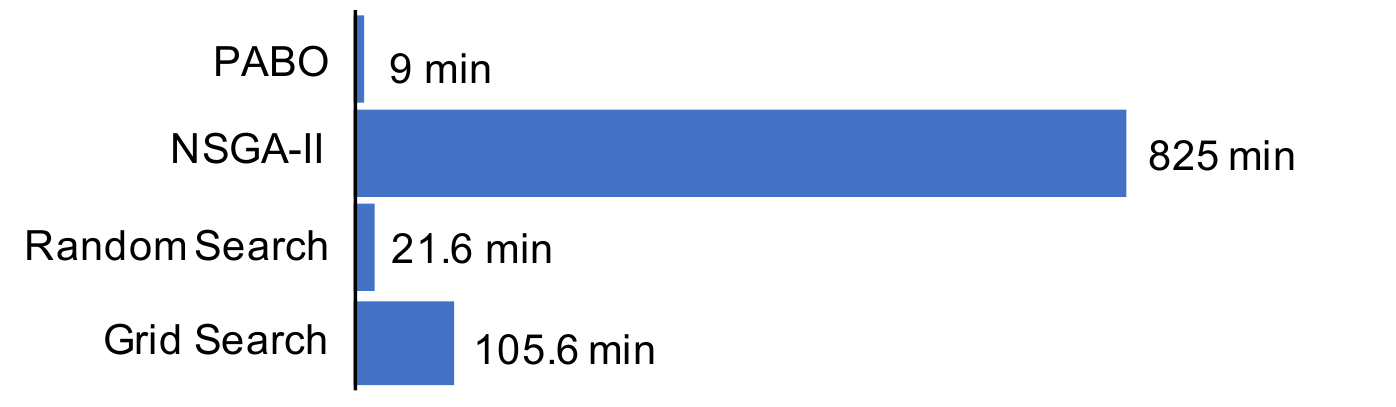}
%\vspace{-0.7cm}
\caption{
Execution time for case study 1. PABO is ~92x faster than state-of-the-art NSGA-II technique}
%\vspace{-0.3cm}
\label{fig_barplot} 
\end{figure}

With only 17 evaluations (out of 192 possible sets of HPs), PABO estimates the Pareto frontier (red dash line in Figure~\ref{fig5}) for the HPs within $\sim$1-2\% percent of the actual Pareto set (gray line in Figure~\ref{fig5}) obtained using the grid search method. 
Compared to the NSGA-II approach, PABO not only estimates Pareto frontier more accurately, but also it is 92x faster. A comparison between the execution time for different techniques is shown in Figure~\ref{fig_barplot}.

In Table~\ref{Tab2}, to further illustrate the impact of HPs, we summarized them for the points A, B, C and D that are shown in Figure~\ref{fig5}. Point A belongs to the Pareto frontier of the network at which we obtained the optimum DNN performance and hardware energy requirement. Point B corresponds to an HP set with minimized energy requirement for hardware, while producing a sub-optimal DNN design. At point C, HPs result in minimum DNN error but with an inefficient hardware design, and the corresponding HP at point D neither optimizes the DNN performance nor hardware energy consumption. It is clear from Table~\ref{Tab2}, that using a joint optimization approach is indispensable for optimal design of both the DNN and the hardware. Moreover, in this case study, selecting HPs given in point A (from Table~\ref{Tab2}) results in up to 40\% decrease in energy requirements for the memristive crossbar accelerator compared to the case where DNN is not optimized for the hardware architecture design (point E shown in Figure~\ref{fig5}).

\begin{table}[]
\tiny
\centering
\caption{Hyperparameter Analysis for case study 1}
%\vspace{-0.3cm}
\label{Tab2}
\resizebox{\columnwidth}{!}{%}
\begin{tabular}{|c|c c c c|}%{llll}
\hline
\textbf{Hyperparameter} & \textbf{A} & \textbf{B} & \textbf{C} & \textbf{D}\\ \hline
\textbf{Dropout} & 0.5 & 0.4 & 0.5 & 0.4\\ %\hline
\textbf{Learning rate} & 0.001 & 0.001 & 0.001 & 0.001\\ %\hline
\textbf{Momentum} & 0.95 & 0.85 & 0.95 & 0.9\\ %\hline
\textbf{Batch size} & 64 & 64 & 64 & 64\\ %\hline
\textbf{\# of FC layers} & 2 & 2 & 2 & 3\\ %\hline
\textbf{\# of conv. layers} & 4 & 4 & 5 & 5\\ %\hline
\textbf{Kernel size, layer 1} & 5 & 5 & 7 & 7\\ %\hline
\textbf{Kernel size, layer 2} & 3 & 3 & 3 & 5\\ %\hline
\textbf{Kernel size, layer 3} & 3 & 3 & 5 & 5\\ %\hline
\textbf{Kernel size, layer 4} & 3 & 3 & 3 & 3\\ \hline
\end{tabular}
}
%\vspace{-0.5cm}
\end{table}

\begin{figure}[h]
%\vspace{-0.2cm}
\centering
\includegraphics [scale=.3]{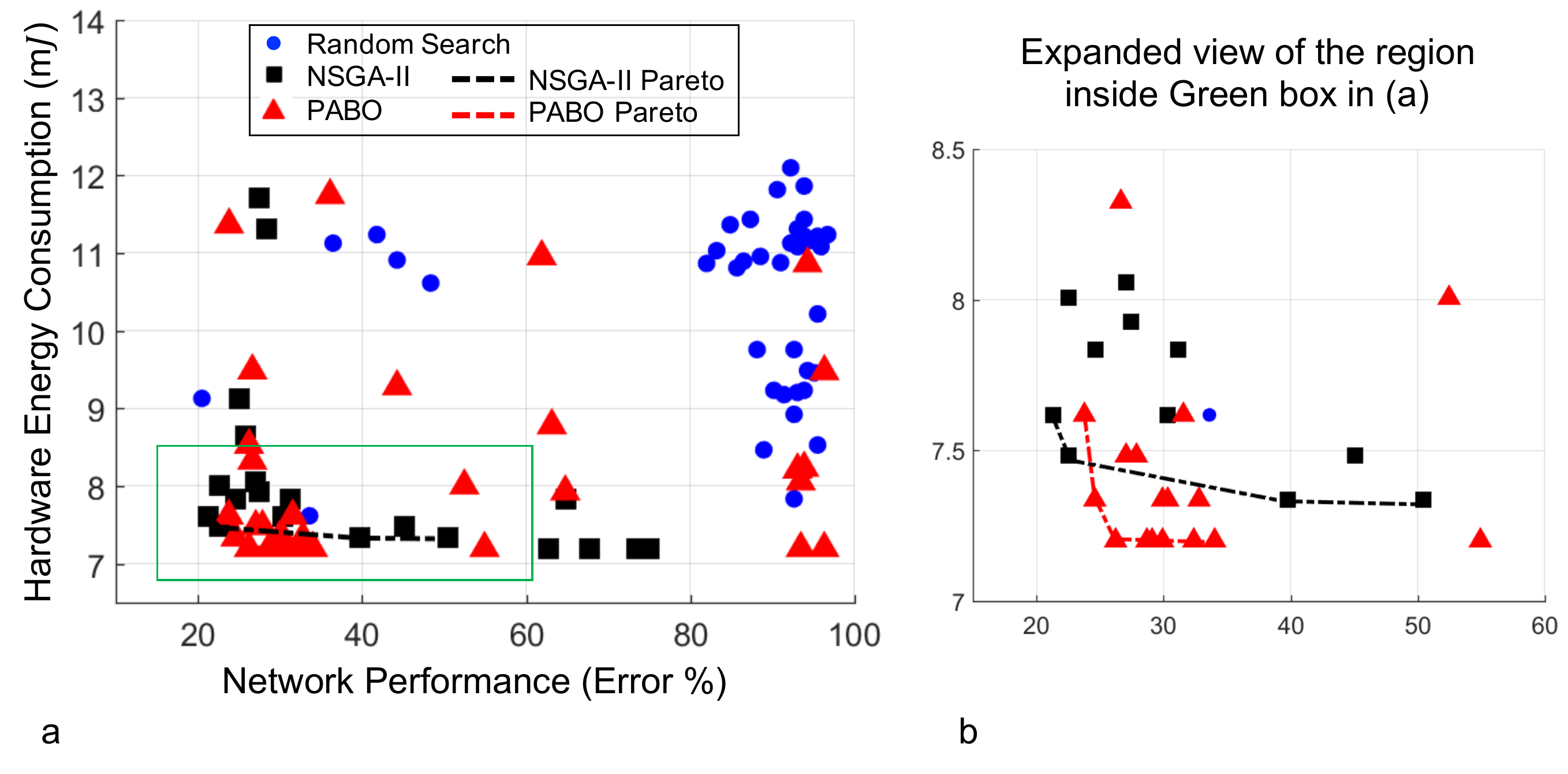}
%\vspace{-0.7cm}
\caption{a. Case study 2: Comparison between PABO, NSGA-II, Grid Search and Random Search for AlexNet netork on Flower17 dataset with 6912 different set of HP combination. 
b. The expanded view of the region inside the green box in panel (a).}
%\vspace{-0.3cm}
\label{fig7_1} 
\end{figure}

\begin{figure}[h]
%\vspace{-0.2cm}
\centering
\includegraphics [scale=.5]{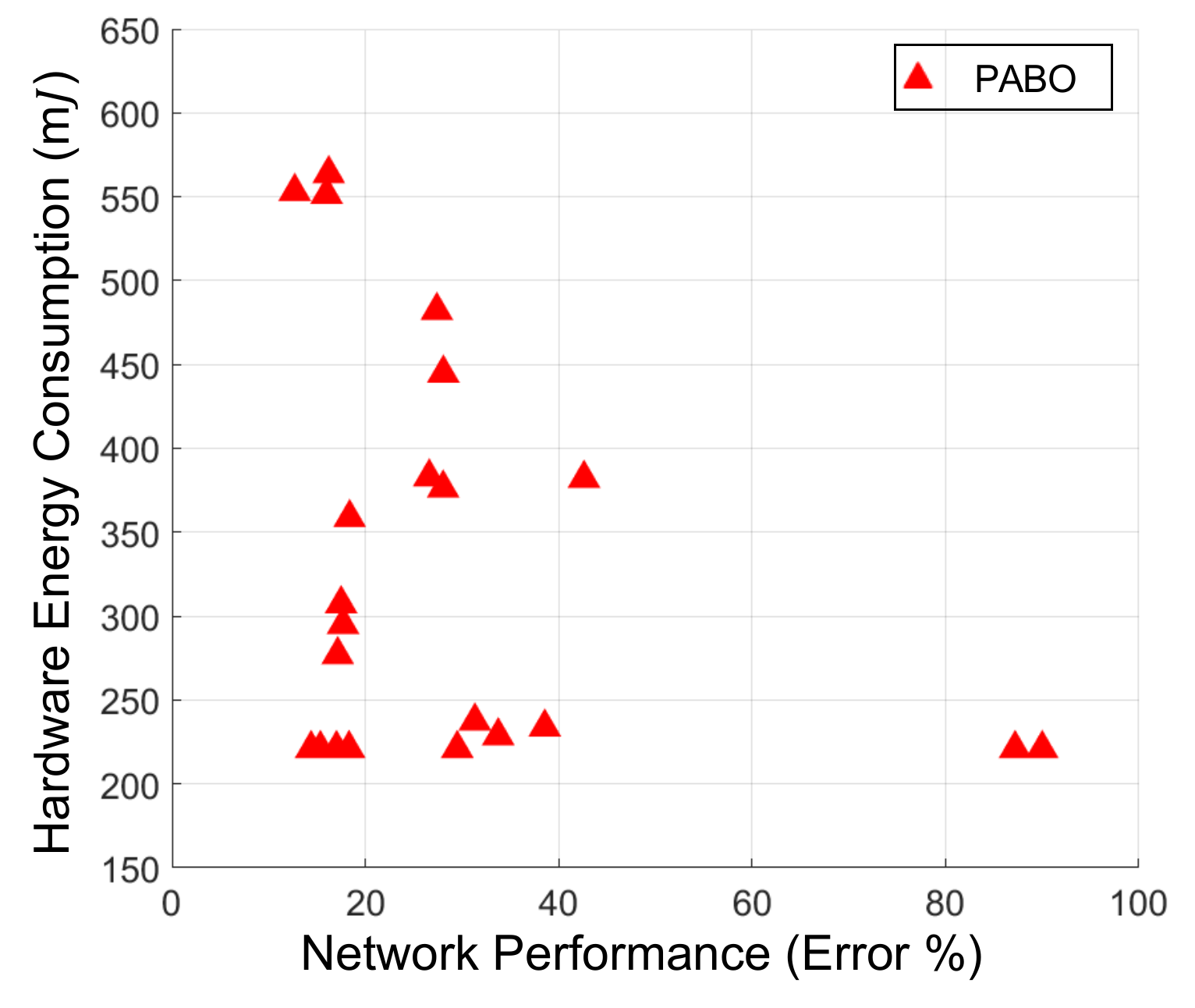}
%\vspace{-0.7cm}
\caption{Case study 3: PABO result for VGG19 network on CIFAR-10 dataset with 3702 different set of HP combination. Other methods cannot be run on this case study due to significant computational requirements. }
\vspace{-0.3cm}
\label{fig7_2} 
\end{figure}

Figures~\ref{fig7_1} and ~\ref{fig7_2} show the results for two additional case studies with more realistic choices of HPs. Details of the HP selections are given in Table~\ref{Tab1}. Figure~\ref{fig7_1} is a case study with 6912 choices of HP combinations on AlexNet architecture with Flower17 dataset. PABO results are shown in red triangles and compared with random search with blue dots and NSGA-II with black squares. Random search is performed with 40 evaluations, NSGA-II had population size of 20 with maximum generation of 100. 
PABO approximates the Pareto frontier with only 33 function evaluations (Training the DNN and computing the hardware energy consumption). Compare to 6000 function evaluations of NSGA-II technique, this leads to $183\times$ faster execution time. 
Note that the results obtained for NSGA-II are the best we could get with the same computational resources we used to perform PABO and other methods. We may improve the NSGA-II's results using supercomputers and significantly longer hours of computation. 

In Figure~\ref{fig7_2}, we used VGG19 network on CIFAR-10 data with 3702 HPs combinations listed in Table 1. Since the network is significantly larger than AlexNet, it is computationally prohibiting to perform NSGA-II using our computational resources. 
On the other hand, PABO was able to estimate the Pareto frontier with only 22 evaluations.

%\vspace{-0.1cm}

\section{Conclusion}

We proposed a novel pseudo agent-based multi-objective hyperparameter optimization technique, deemed PABO, that can maximize the neural network performance while minimizing the energy requirements of the underlying hardware. 
PABO uses Bayesian optimization with Gaussian processes and acquisition function along with a supervisor agent, to estimate the Pareto frontier that shows the optimum HP sets for maximum neural network performance and minimum hardware energy consumption.

We tested PABO on both AlexNet and VGG19 neural network while designing a memristive crossbar accelerator, and compared it with other algorithms.
Superior performance of our method both in terms of accuracy and computational time was demonstrated. $100$x increase in computational speed compared to the NSGA-II algorithm was demonstrated. 
It is important to note that PABO is not limited to a specific neural network or hardware architecture, and can be applied to multiple (more than two) black-box functions corresponding to different design requirements.

%\section{Acknowledgement}
%This work was supported in part by the Center for Brain-inspired Computing Enabling Autonomous Intelligence (C-BRIC), one of six centers in JUMP, a Semiconductor Research Corporation (SRC) program sponsored by DARPA, in part by Intel/Semiconductor Research Corporation Education Alliance (SCRCEA), and in part by the Vannevar Bush Faculty Fellowship.
%\vspace{-0.1cm}
\renewcommand*{\bibfont}{\normalsize}
\bibliographystyle{IEEEbib}%ACM-Reference-Format}
\bibliography{MP_Abstract}%Refs_test}%sample-bibliography}
\end{document}